\pdfoutput=1

\documentclass[11pt]{article}

\usepackage{acl}

\usepackage{times}
\usepackage{latexsym}

\usepackage[T1]{fontenc}

\usepackage[utf8]{inputenc}

\usepackage{microtype}
\usepackage{amsmath}
\usepackage{amssymb}
\usepackage{graphicx}
\usepackage{subcaption}
\usepackage{pgfplots}
\usetikzlibrary{patterns}
\usepackage{filecontents}
\usepackage{algorithmic,algorithm}
\usepackage{csvsimple}
\usepackage{xspace}
\usepackage{mathtools}

\usepackage{siunitx}
\usepackage{array}
\usepackage{booktabs}
\usepackage{caption}
\usepackage{enumitem}

\usepackage{pifont}
\newcommand{\cmark}{\ding{51}}
\newcommand{\xmark}{\ding{55}}
\newcommand\independent{\protect\mathpalette{\protect\independenT}{\perp}}
\def\independenT#1#2{\mathrel{\rlap{$#1#2$}\mkern2mu{#1#2}}}

\definecolor{blue}{HTML}{4285f4}
\definecolor{lightblue}{HTML}{c9daf8}
\definecolor{darkblue}{HTML}{1c4587}
\definecolor{green}{HTML}{34a853}
\definecolor{lightgreen}{HTML}{d9ead3}
\definecolor{orange}{HTML}{ff9900}
\definecolor{lightorange}{HTML}{fce5cd}
\definecolor{lightred}{HTML}{e06666}
\definecolor{purple}{HTML}{9900ff}
\definecolor{lightpurple}{HTML}{b4a7d6}
\definecolor{gray}{HTML}{cccccc}

\definecolor{sqcolor}{HTML}{4285f4} %
\definecolor{nqcolor}{HTML}{0b5394} %
\definecolor{hpcolor}{HTML}{9fc5e8} %
\definecolor{trvcolor}{HTML}{ff9900} %
\definecolor{seacolor}{HTML}{f9cb9c} %
\definecolor{newscolor}{HTML}{cc4125} %

\newcommand{\benchmark}[1]{\textsc{#1}\xspace}
\newcommand{\squad}{\benchmark{SQuAD}}
\newcommand{\newsqa}{NewsQA\xspace}
\newcommand{\hotpotqa}{HotpotQA\xspace}
\newcommand{\nqshort}{NQ\xspace}
\newcommand{\triviaqa}{TriviaQA\xspace}
\newcommand{\searchqa}{SearchQA\xspace}

\definecolor{brandeisblue}{rgb}{0.0, 0.44, 1.0}
\newcommand{\ggao}[1]{{\color{brandeisblue}\{\textit{#1}\}$_{ge}$}}
\newcommand{\td}[1]{{\color{lightred}\{\textit{#1}\}$_{TODO}$}}

\newcommand{\eat}[1]{}

\newcommand{\reals}{{\rm I\!R}}

\newcommand{\question}{\bar{q}}
\newcommand{\questiontoken}{q}
\newcommand{\context}{\bar{c}}
\newcommand{\contexttoken}{c}
\newcommand{\answer}{y}
\newcommand{\predictedanswer}{\hat{\answer}}

\newcommand{\qapolicy}{\pi}
\newcommand{\reward}{r}
\newcommand{\regret}{R}
\newcommand{\timestep}{t}
\newcommand{\params}{\theta}

\newcommand\cn{$^\diamondsuit$}
\newcommand\ut{$^\clubsuit$}

\title{Simulating Bandit Learning from User Feedback \\ for Extractive Question Answering}

\author{Ge Gao\cn, Eunsol Choi\ut\phantom{,} \textnormal{and} Yoav Artzi\cn \\
\cn Department of Computer Science and Cornell Tech, Cornell University\\ 
\ut Department of Computer Science, The University of Texas at Austin \\
{\tt ggao@cs.cornell.edu} \hspace{0.5em} {\tt eunsol@utexas.edu} \hspace{0.5em} {\tt yoav@cs.cornell.edu}\\}

\begin{document}
\maketitle
\begin{abstract}

We study learning from user feedback for extractive question answering by simulating feedback using supervised data. We cast the problem as contextual bandit learning, and analyze the characteristics of several learning scenarios with focus on reducing data annotation. We show that systems initially trained on a small number of examples can dramatically improve given feedback from users on model-predicted answers, and that one can use existing datasets to deploy systems in new domains without any annotation, but instead improving the system on-the-fly via user feedback.

\end{abstract}

\section{Introduction}\label{sec:intro}

Explicit feedback from users of NLP systems can be used to continually improve system performance. 
For example, a user posing a question to a question-answering (QA) system can mark if a predicted phrase is a valid answer given the context from which it was extracted. 
However, the dominant paradigm in NLP separates model training from deployment, leaving models static following learning and throughout interaction with users. 
This approach misses opportunities for learning during system usage, which beside several exceptions we discuss in \autoref{sec:related} is understudied in NLP. 
In this paper, we study the potential of learning from explicit user feedback for extractive QA through simulation studies.

\begin{figure}[t!]
\centering
\includegraphics[width=0.45\textwidth]{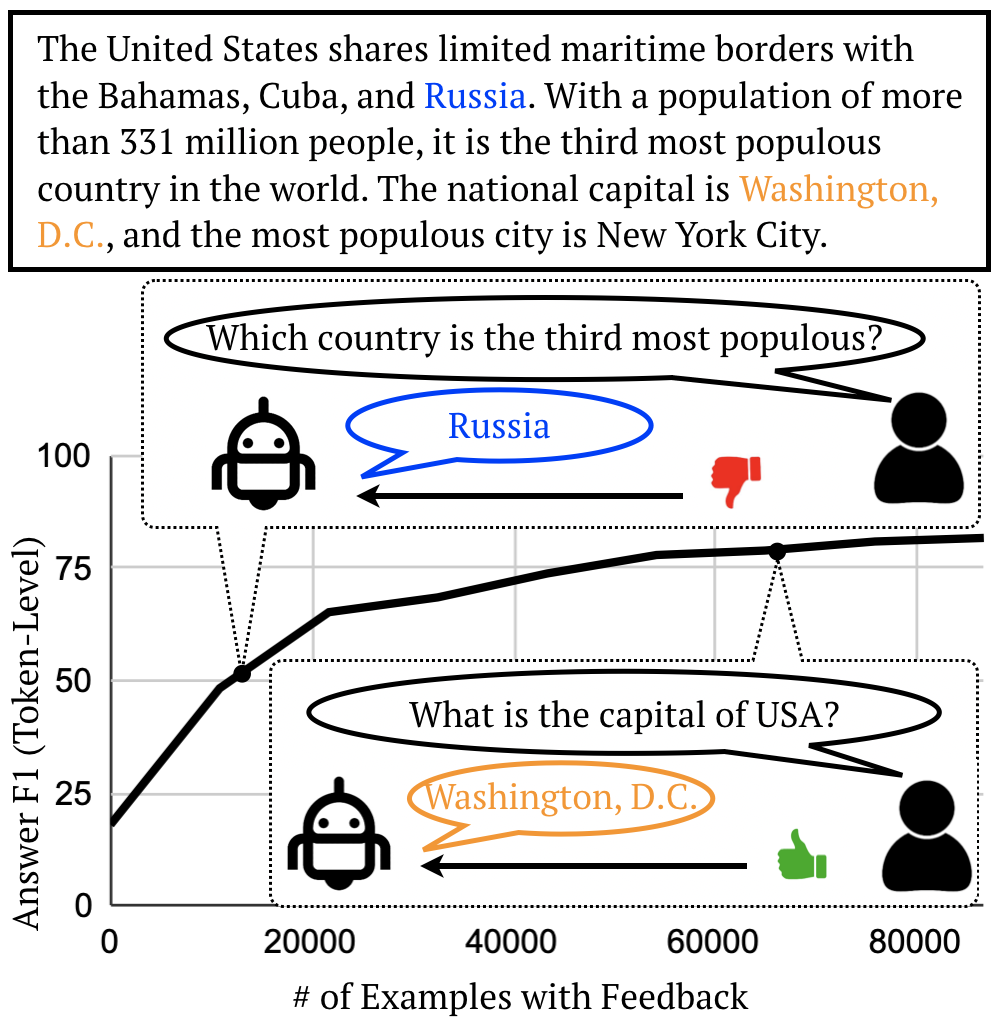}
\caption{Illustration of an interaction setup for learning from user feedback for QA, and its potential. Given a user question, the system outputs an answer and highlights it in its context. The user validates the answer given the context with binary feedback. 
We show performance progression from one of our online learning experiments on \squad with hand-crafted illustrative examples at two time steps.}\label{fig:ex}
\end{figure}

Extractive QA is a popular testbed for language reasoning, with rich prior work on datasets~\cite[e.g.,][]{Rajpurkar2016SQuAD1Q}, task design~\cite{Yang2018HotpotQAAD,Choi2018QuACQA}, and model architecture development~\cite{Seo2017BidirectionalAF,Yu2018QANetCL}. 
Learning from interaction with users remains relatively understudied, even though QA is well positioned to elicit user feedback. 
An extracted answer can be clearly visualized within its supporting context, and a language-proficient user can then easily validate if the answer is supported or not.\footnote{Answers could also come from erroneous or deceitful contexts. This important problem is not studied by most work in extractive QA, including ours. We leave it for future work.}
This allows for simple binary feedback, and creates a contextual bandit learning scenario~\cite{Auer2002:nonstochastic-bandit,Langford2007TheEA}. 
Figure~\ref{fig:ex} illustrates this learning signal and its potential.

We simulate user feedback using several widely used QA datasets, and use it as a bandit signal for learning. 
We study the empirical characteristics of the learning process, including its performance, sensitivity to initial system performance, and trade-offs between online and offline learning. 
We also simulate zero-annotation domain adaptation, where we deploy a QA system trained from supervised data in one domain and adapt it solely from user feedback in a new domain. %

This learning scenario can mitigate fundamental problems in extractive QA. 
It reduces data collection costs, by delegating much of the learning to interaction with users. 
It can avoid data collection artifacts because the data comes from the  actual system deployment, unlike data from an annotation effort that often involves design decisions immaterial to the system's use case. 
For example, sharing question- and answer-annotator roles~\cite{Rajpurkar2016SQuAD1Q}, which is detrimental to emulate information seeking behavior~\cite{Choi2018QuACQA}.  
Finally, it gives systems the potential to evolve over time as the world changes~\cite{Lazaridou2021:temporal-generalization,Zhang2021:situatedqa}.

Our simulation experiments show that user feedback is an effective signal to continually improve QA systems across multiple benchmarks. 
For example, an initial system trained with a small amount of \squad~\cite{Rajpurkar2016SQuAD1Q} annotations (64 examples) improves from 18 to 81.6 F1 score, and adapting a \searchqa~\cite{Dunn2017SearchQAAN} system to \squad through user feedback improves it from 45 to 84 F1 score. 
Our study shows the impact of initial system performance, trade-offs between online and offline learning, and the impact of source domain on adaptation. 
These results create the base for future work that goes beyond simulation to use feedback from human users to improve extractive QA systems. 
Our code is publicly available at \url{https://github.com/lil-lab/bandit-qa}.

\section{Learning and Interaction Scenario}\label{sec:scenario}

We study a scenario where a QA model learns from explicit user feedback. We formulate learning as a contextual bandit problem. 
The input to the learner is a question-context pair, where the context paragraph contains the answer to the question. %
The output is a single span in the context paragraph that is the answer to the question. 

Given a question-context pair, the model predicts an answer span. The  user then provides feedback about the model's predicted answer, which is used to update the model parameters.
We intentionally experiment with simple binary feedback and basic learning algorithms, to provide a baseline for what more advanced methods could achieve with as few assumptions as possible.  

\paragraph{Background: Contextual Bandit Learning}
In a stochastic (i.i.d.) contextual bandit learning problem, at each time step $t$, the learner independently observes a context\footnote{The term \emph{context} here refers to the input to the learner policy, and is different from the term \emph{context} as we use it later in extractive QA, where the term \emph{context} refers to the evidence document given as input to the model.} $x^{(t)} \sim D$ sampled from the data distribution $D$, chooses an action $\answer^{(t)}$ according to a policy $\qapolicy$, and observes a reward $\reward^{(t)} \in \mathbb{R}$. The learner only observes the reward $\reward^{(t)}$ corresponding to the chosen action $\answer^{(t)}$. 
The learner aims to minimize the cumulative regret. 
Intuitively, regret is the deficit suffered by the learner relative to the optimal policy up to a specific time step. 
Formally, the cumulative regret at time $T$ is computed with respect to the optimal policy $\qapolicy^* \in \arg\max_{\qapolicy \in \Pi} \mathbb{E}_{(x, \answer, \reward) \sim (D, \qapolicy)}[\reward]$: 
\begin{equation}
    \regret_T \coloneqq \sum_{t=1}^T \reward^{*(t)} - \sum_{t=1}^T \reward^{(t)}\;\;,\label{eq:banditregret}
\end{equation}
where $\Pi$ is the set of all policies, $\reward^{(t)}$ is the reward observed at time $t$ and $\reward^{*(t)}$ is the reward that the optimal policy $\qapolicy^*$ would observe. Minimising the cumulative regret is equivalent to maximising the total reward.\footnote{Equivalently, the problem is often formulated as loss minimization~\cite{Bietti2018ACB}.} 
A key challenge in contextual bandit learning is to balance exploration and exploitation to minimize overall regret.

\paragraph{Scenario Formulation}
Let a question $\question$ be a sequence of $m$ tokens $\langle \questiontoken_1,\dots,\questiontoken_m\rangle$ and a context paragraph $\context$ be a sequence of $n$ tokens $\langle \contexttoken_1,\dots,\contexttoken_n\rangle$. 
An extractive QA model\footnote{In bandit literature, the term \emph{policy} is more commonly used. We use the term \emph{model} from here on to align with the QA literature.} $\qapolicy$ predicts a span $\predictedanswer = \langle \contexttoken_i,\dots,\contexttoken_j\rangle$ where $i, j \in [1, n]$ and $i \leq j$ in the context $\context$ as an answer. 
When relevant, we denote $\qapolicy_\params$ as a QA model parameterized by $\params$.

We formalize learning as a contextual bandit process: at each time step $\timestep$, the model is given a question-context pair $(\question^{(t)}, \context^{(t)})$, predicts an answer span $\predictedanswer$, and receives a reward $\reward^{(t)} \in \reals$. 
The learner's goal is to maximize the total reward $\sum_{t=1}^T \reward^{(t)}$.
This formulation reflects a setup where, given a question-context pair, the QA system interacts with a user, who validates the model-predicted answer in context, and provides feedback which is mapped to numerical reward.

\paragraph{Learning Algorithm}

We learn using policy gradient. 
Our learner is similar to REINFORCE~\cite{Sutton1998:rl-book-second,Williams2004SimpleSG}, but we use $\arg\max$ to predict answers instead of Monte Carlo sampling from the model's output distribution.\footnote{Early experiments showed that sampling is not as beneficial as $\arg\max$, potentially because of the relatively large output space of extractive QA. \citet{Yao2020:imitation-sempar} made a similar observation for semantic parsing, and  \citet{Lawrence2017CounterfactualLF} used $\arg\max$ predictions for bandit learning in statistical machine translation. \autoref{tab:sampling} in \autoref{sec:discussion} provides our experimental results with sampling.}

We study online and offline learning, also referred to as on- and off-policy. 
In online learning (\autoref{alg:online}), the model identity is maintained between prediction and update; the parameter values that are updated are the same that were used to generate the output receiving reward. 
This entails that a reward is only used once, to update the model after observing it. 
In offline learning (\autoref{alg:offline}), this relation between update and prediction does not hold. 
The learner observes reward, often across many examples, and may use it to update the model many times, even after the parameters drifted arbitrarily far from these that generated the prediction. 
In practice, we observe reward for the entire length of the simulation ($T$ steps) and then update for $E$ epochs. The reward is re-weighted to provide an unbiased estimation using inverse propensity score~\cite[IPS;][]{Horvitz1952AGO}. We clip the debiasing coefficient to avoid amplifying examples with large coefficients (line~\ref{alg:offline:ips}, \autoref{alg:offline}).

\begin{algorithm}[t]
\caption{Online learning.}
\label{alg:online}
\begin{algorithmic}[1]
  \small
  \FOR{$t = 1 \cdots$}
      \STATE Receive a question $\question^{(t)}$ and context $\context^{(k)}$
      \STATE Predict an answer $\predictedanswer^{(t)} \gets \arg\max_\answer \qapolicy_{\params}(\answer \mid \question^{(t)}, \context^{(t)})$ \vspace{-9pt}
      \STATE Observe a reward $\reward^{(t)}$
      \STATE Update the model parameters $\theta$ using the gradient \\ \hspace*{1cm}$r^{(t)}\nabla_{\params}\log \qapolicy_{\params}(\predictedanswer^{(t)} \mid \question^{(t)}, \context^{(t)})$
  \ENDFOR
\end{algorithmic}
\end{algorithm}

\begin{algorithm}[t]
  \caption{Offline learning.}
  \label{alg:offline}
\begin{algorithmic}[1]
  \small
  \FOR{$t = 1 \cdots T$}
      \STATE Receive a question $\question^{(t)}$ and context $\context^{(t)}$
      \STATE Predict an answer $\predictedanswer^{(t)} \gets \arg\max_\answer\qapolicy_{\params}(\answer \mid \question^{(t)}, \context^{(t)})$ 
      \vspace{-9pt}
      \STATE $p^{(t)} \gets \qapolicy_{\params}(\predictedanswer^{(t)} \mid \question^{(t)}, \context^{(t)})$
      \STATE Observe a reward $\reward^{(t)}$
  \ENDFOR
  \FOR{$E$ epochs}
  \FOR{$t = 1 \cdots T$}
     \STATE Compute clipped importance-weighted reward according to the current model parameters: 
     \STATE $\reward' \gets {\rm clip}(\frac{\qapolicy_{\params}(\predictedanswer^{(t)} \mid \question^{(t)}, \context^{(t)})}{p^{(t)}},0,1)\reward^{(t)}$\label{alg:offline:ips}
      \STATE Update the model parameters $\theta$ using the gradient \\ \hspace*{1cm}$r'\nabla_{\params}\log \qapolicy_{\params}(\predictedanswer^{(t)} \mid \question^{(t)}, \context^{(t)})$
  \ENDFOR
  \ENDFOR
\end{algorithmic}
\end{algorithm}

In general, offline learning is easier to implement because updating the model is not integrated with its deployment. Offline learning also uses a training loop that is similar to optimization practices in supervised learning. This allows to iterate over the data multiple times, albeit with the same feedback signal on each example. 
However, online learning often has lower regret as the model is updated after each interaction.
It may also lead to higher overall performance, because as the model improves early on, it may observe more positive feedback overall, which is generally more informative. 
We empirically study these trade-offs in \autoref{sec:online-learning} and~\ref{sec:offline-learning}.

\paragraph{Evaluating Performance}
We evaluate model performance using token-level F1 on a held-out test set, as commonly done in the QA literature~\cite{Rajpurkar2016SQuAD1Q}. 
We also estimate the learner regret (\autoref{eq:banditregret}).
Computing regret requires access to the an oracle $\qapolicy^*$. We use human annotation as an estimate (\autoref{sec:sim-setup}).\footnote{Our oracle is an estimate because of annotation noise and ambiguity in exact span selection.}

\paragraph{Comparison to Supervised Learning}

In supervised learning, the data distribution is not dependent on the model, but on a fixed training set $\{(\question^{(t)}, \context^{(t)}, \answer^{(t)}) \}_{t=1}^T$.  
In contrast, bandit learners are provided with reward data that depends on the model itself: $\{(\question^{(t)}, \context^{(t)}, \predictedanswer^{(t)}, \reward^{(t)}) \}_{t=1}^T$ where $\reward$ is the reward for the model prediction $\predictedanswer^{(t)} = \arg\max_\answer \qapolicy_{\params}(\answer \mid \question^{(t)}, \context^{(t)})$ at time step $t$. 
Such feedback can be freely gathered from users interacting with the model, while building supervised datasets requires costly annotation. 
This learning signal can also reflect changing task properties (e.g., world changes) to allow systems to adapt, and its origin in the deployed system use makes it more robust to biases introduced during annotation.

\section{Simulation Setup} \label{sec:sim-setup}

We initialize our model with supervised data, and then simulate bandit feedback using supervised data annotations. Initialization is critical so the model does not return random answers, which are likely to be all bad because of the large output space. 
We use relatively little supervised data from the same domain for in-domain experiments (\autoref{sec:online-learning} and~\ref{sec:offline-learning}) to focus on the data annotation reduction potential of user feedback. 
For domain adaptation, we assume access to a large amount of training data in the source domain, and no annotated data in the target domain (\autoref{sec:domain-adaptation}).

\paragraph{Reward} 

We use supervised data annotations to simulate the reward. 
If the predicted answer span is an exact match index-wise to the annotated span, the learner observes a positive reward of 1.0, and a negative reward of -0.1 otherwise.\footnote{We experimented with other reward values, but did not observe a significant difference in performance (\autoref{sec:discussion}).} 
This reward signal is stricter than QA evaluation metrics (token-level F1 or exact match after normalization).\footnote{Normalization includes lowercasing, modifying spacing, removing articles and punctuation, etc. NaturalQuestions~\cite[\nqshort;][]{Kwiatkowski2019NaturalQA} is an exception, with an exact index match measure that has similar strictness.}

\paragraph{Noise Simulation}
We study robustness by simulating noisy feedback via reward perturbation: randomly flipping the binary reward with a fixed probability of 8\% or 20\% as the noise ratio.\footnote{Even without our noise simulation, the simulated feedback inherits the noise from the annotation, either from crowdsourcing or distant supervision.}

\section{Experimental Setup}\label{sec:exp-setup}

\paragraph{Data}

We use six English QA datasets that provide substantial amount of annotated training data taken from the MRQA training portion~\cite{fisch2019mrqa}: \squad~\cite{Rajpurkar2016SQuAD1Q}, \newsqa~\cite{Trischler2017NewsQAAM}, \searchqa~\cite{Dunn2017SearchQAAN}, \triviaqa~\cite{Joshi2017TriviaQAAL}, \hotpotqa~\cite{Yang2018HotpotQAAD}, and NaturalQuestions~\cite[\nqshort;][]{Kwiatkowski2019NaturalQA}. The MRQA benchmark simplifies all datasets so that each example has a single span answer with a limited evidence document length (truncated at 800 tokens).
\autoref{tab:data} in \autoref{sec:extra} provides dataset details.
We compute performance measures and learning curves on development sets following prior work~\cite{Rajpurkar2016SQuAD1Q, Ram2021FewShotQA}. 

\paragraph{Model}
We conduct experiments with a pretrained SpanBERT model~\cite{Joshi2020SpanBERTIP}. We fine-tune the pre-trained SpanBERT-base model during initial learning and our simulations.

\paragraph{Implementation Details} 
We use Hugging Face Transformers~\cite{Wolf2019HuggingFacesTS}. 
When training initial models with little in-domain supervised data (\autoref{sec:online-learning}; \autoref{sec:offline-learning}), we use a learning rate of 3e-5 with a linear schedule, batch size 10, and 10 epochs. We obtain the sets of 64, 256, or 1{,}024 examples from prior work~\cite{Ram2021FewShotQA}.\footnote{We use the seed 46 sets publicly available at \url{https://github.com/oriram/splinter}.} 
For models initially trained on complete datasets (\autoref{sec:domain-adaptation}), we use a learning rate 2e-5 with a linear schedule, batch size 40, and 4 epochs. 

In simulation experiments, we use batch size 40. We turn off dropout to simulate interaction with users in deployment. For single-pass online learning experiments  (\autoref{sec:online-learning}; \autoref{sec:domain-adaptation}), we use a constant learning rate of 1e-5. 
For offline learning experiments (\autoref{sec:offline-learning}), we train the model for 3 epochs on the collected feedback with a linear schedule learning rate of 3e-5. 

Online experiments with \squad, \hotpotqa, \nqshort, and \newsqa take 2--4h each on one NVIDIA GeForce RTX 2080 Ti; 2.5--6h for offline. 
For \triviaqa and \searchqa, each online simulation experiment on one NVIDIA TITAN RTX takes 4--9.5h; 9--20h for offline.

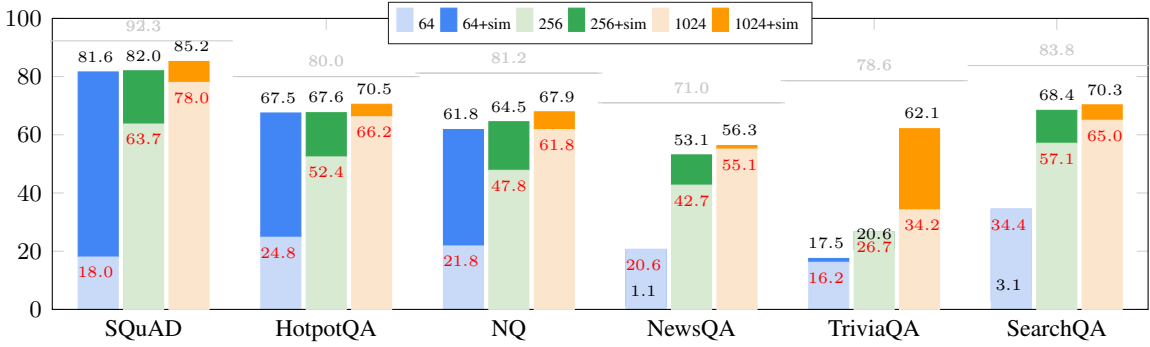
\begin{figure*}[t!] \vspace{-10pt}
\centering
\begin{tikzpicture}[
  /pgfplots/every axis/.style={
    ybar stacked,
    ymin=0,ymax=100,
    x tick label style={rotate=0,anchor=north},
    symbolic x coords={SQuAD, HotpotQA, NQ, NewsQA, TriviaQA, SearchQA},
  bar width=15,
  width=\textwidth,
  height=.34\textwidth,
  xtick=data, 
  font=\small,
  nodes near coords,
  },
]

\begin{axis}[hide axis, bar shift=-17, nodes near coords style={font=\tiny, xshift=-18, /pgf/number format/fixed zerofill, /pgf/number format/precision=1}]
    \addplot [lightblue, fill=lightblue, nodes near coords align={anchor=north,color=red}] table [x=dataset, y=64, col sep=comma] {in-online-pfm.csv};
    \addplot [blue, fill=blue, nodes near coords align={anchor=south,color=black}] table [x=dataset, y=64diff, col sep=comma] {in-online-pfm.csv};
\end{axis}

\begin{axis}[nodes near coords,
    nodes near coords style={font=\tiny, /pgf/number format/fixed zerofill, /pgf/number format/precision=1}]
    \addplot [lightgreen, fill=lightgreen, nodes near coords align={anchor=north, color=red}] table [x=dataset, y=256, col sep=comma] {in-online-pfm.csv};
    \addplot [green, fill=green, nodes near coords align={anchor=south,color=black}, plot graphics/node/.style={text=red}] table [x=dataset, y=256diff, col sep=comma] {in-online-pfm.csv};
\end{axis}

\begin{axis}[hide axis, bar shift=17, nodes near coords,
    nodes near coords style={font=\tiny,xshift=18, /pgf/number format/fixed zerofill, /pgf/number format/precision=1}]
    \addplot [lightorange, fill=lightorange, nodes near coords align={anchor=north,color=red}] table [x=dataset, y=1024, col sep=comma] {in-online-pfm.csv};
    \addplot [orange, fill=orange, nodes near coords align={anchor=south,color=black}] table [x=dataset, y=1024diff, col sep=comma] {in-online-pfm.csv};
\end{axis}

\begin{axis}[hide axis, clip=false, xshift=-35, nodes near coords,
    nodes near coords style={font=\tiny, color=gray, /pgf/number format/fixed zerofill, /pgf/number format/precision=1}]
    \addplot [gray, jump mark left, nodes near coords align={anchor=south, xshift=35}] table [x=dataset, y=supervised, col sep=comma] {in-online-pfm.csv};
\end{axis}
\begin{axis}[hide axis, clip=false, xshift=35, nodes near coords, nodes near coords style={font=\tiny, color=gray, /pgf/number format/fixed zerofill, /pgf/number format/precision=1}]
    \addplot [gray, jump mark right, nodes near coords align={anchor=south, xshift=-35}] table [x=dataset, y=supervised, col sep=comma] {in-online-pfm.csv};
\end{axis}

\begin{axis}[hide axis, xmin=SQuAD,xmax=SearchQA,ymin=0,ymax=100,height=.1\textwidth,
    legend style={draw=white!15!black,at={(0.5,190)},anchor=south},legend columns=-1,font=\tiny,
    ]
        \addlegendimage{lightblue,fill=lightblue}
        \addlegendentry{64};
        \addlegendimage{blue,fill=blue}
        \addlegendentry{64+sim};
        \addlegendimage{lightgreen, fill=lightgreen}
        \addlegendentry{256};
        \addlegendimage{green, fill=green}
        \addlegendentry{256+sim};
        \addlegendimage{lightorange, fill=lightorange}
        \addlegendentry{1024};
        \addlegendimage{orange,fill=orange}
        \addlegendentry{1024+sim};
\end{axis}
\end{tikzpicture}\vspace{-5pt}
\caption{Online in-domain simulation development F1 performance. Horizontal grey lines represent the supervised training performance on each dataset. Data labels in red are performance of initial models trained on 64, 256, or 1024 examples (i.e., lighter bars). Darker bars and black data labels represent simulation performance. Lower simulation performance (e.g., \newsqa 64+sim) indicate degradation in performance following simulation.}
\label{fig:in-pfm-online}
\end{figure*}

\begin{figure*}[t!]\vspace{-5pt}
\centering
\begin{tabular}{p{.27\textwidth} p{.27\textwidth} p{.27\textwidth}}
\multicolumn{3}{c}{\begin{tikzpicture} 
    \begin{axis}[hide axis, xmin=10,xmax=50,ymin=0,ymax=0.4,height=.1\textwidth,
    legend style={draw=white!15!black,legend cell align=left},legend columns=-1,font=\tiny,
    ]
        \addlegendimage{blue,solid,thick}
        \addlegendentry{64};
        \addlegendimage{blue,densely dashed,thick}
        \addlegendentry{64 w .08};
        \addlegendimage{blue,densely dotted,thick}
        \addlegendentry{64 w .2};
        \addlegendimage{green,solid,thick}
        \addlegendentry{256};
        \addlegendimage{green,densely dashed,thick}
        \addlegendentry{256 w .08};
        \addlegendimage{green,densely dotted,thick}
        \addlegendentry{256 w .2};
        \addlegendimage{orange,solid,thick}
        \addlegendentry{1024};
        \addlegendimage{orange,densely dashed,thick}
        \addlegendentry{1024 w .08};
        \addlegendimage{orange,densely dotted,thick}
        \addlegendentry{1024 w .2};
    \end{axis}
\end{tikzpicture}} \\
\begin{tikzpicture}
    \begin{axis}[width=.35\textwidth, xmin=0,xmax=117400,xtick={0,25000,50000,75000,100000}, xlabel=SQuAD,ymin=0,ymax=90,grid=major,font=\small]
        \addplot[blue,solid,thick,mark=none] table [x=a, y=b, col sep=comma] {in-online-sq.csv};
        \addplot[blue,densely dashed,thick,mark=none] table [x=a, y=c, col sep=comma] {in-online-sq.csv};
        \addplot[blue,densely dotted,thick,mark=none] table [x=a, y=d, col sep=comma] {in-online-sq.csv};
        \addplot[green,solid,thick,mark=none] table [x=a, y=e, col sep=comma] {in-online-sq.csv};
        \addplot[green,densely dashed,thick,mark=none] table [x=a, y=f, col sep=comma] {in-online-sq.csv};
        \addplot[green,densely dotted,thick,mark=none] table [x=a, y=g, col sep=comma] {in-online-sq.csv};
        \addplot[orange,solid,thick,mark=none] table [x=a, y=h, col sep=comma] {in-online-sq.csv};
        \addplot[orange,densely dashed,thick,mark=none] table [x=a, y=i, col sep=comma] {in-online-sq.csv};
        \addplot[orange,densely dotted,thick,mark=none] table [x=a, y=j, col sep=comma] {in-online-sq.csv};
    \end{axis}
\end{tikzpicture} &
\begin{tikzpicture}
\begin{axis}[width=.35\textwidth, xmin=0,xmax=117400,xtick={0,25000,50000,75000,100000}, xlabel=HotpotQA,ymin=0,ymax=90,grid=major,font=\small]
\addplot[blue,solid,thick,mark=none] table [x=a, y=b, col sep=comma] {in-online-hp.csv};
\addplot[blue,densely dashed,thick,mark=none] table [x=a, y=c, col sep=comma] {in-online-hp.csv};
\addplot[blue,densely dotted,thick,mark=none] table [x=a, y=d, col sep=comma] {in-online-hp.csv};
\addplot[green,solid,thick,mark=none] table [x=a, y=e, col sep=comma] {in-online-hp.csv};
\addplot[green,densely dashed,thick,mark=none] table [x=a, y=f, col sep=comma] {in-online-hp.csv};
\addplot[green,densely dotted,thick,mark=none] table [x=a, y=g, col sep=comma] {in-online-hp.csv};
\addplot[orange,solid,thick,mark=none] table [x=a, y=h, col sep=comma] {in-online-hp.csv};
\addplot[orange,densely dashed,thick,mark=none] table [x=a, y=i, col sep=comma] {in-online-hp.csv};
\addplot[orange,densely dotted,thick,mark=none] table [x=a, y=j, col sep=comma] {in-online-hp.csv};
\end{axis}
\end{tikzpicture} &
\begin{tikzpicture}
\begin{axis}[width=.35\textwidth, xmin=0,xmax=117400,xtick={0,25000,50000,75000,100000}, xlabel=NQ,ymin=0,ymax=90,grid=major,font=\small]
\addplot[blue,solid,thick,mark=none] table [x=a, y=b, col sep=comma] {in-online-nq.csv};
\addplot[blue,densely dashed,thick,mark=none] table [x=a, y=c, col sep=comma] {in-online-nq.csv};
\addplot[blue,densely dotted,thick,mark=none] table [x=a, y=d, col sep=comma] {in-online-nq.csv};
\addplot[green,solid,thick,mark=none] table [x=a, y=e, col sep=comma] {in-online-nq.csv};
\addplot[green,densely dashed,thick,mark=none] table [x=a, y=f, col sep=comma] {in-online-nq.csv};
\addplot[green,densely dotted,thick,mark=none] table [x=a, y=g, col sep=comma] {in-online-nq.csv};
\addplot[orange,solid,thick,mark=none] table [x=a, y=h, col sep=comma] {in-online-nq.csv};
\addplot[orange,densely dashed,thick,mark=none] table [x=a, y=i, col sep=comma] {in-online-nq.csv};
\addplot[orange,densely dotted,thick,mark=none] table [x=a, y=j, col sep=comma] {in-online-nq.csv};
\end{axis}
\end{tikzpicture} \\
\begin{tikzpicture}
\begin{axis}[width=.35\textwidth, xmin=0,xmax=117400,xtick={0,25000,50000,75000,100000}, xlabel=NewsQA,ymin=0,ymax=90,grid=major,font=\small]
\addplot[blue,solid,thick,mark=none] table [x=a, y=b, col sep=comma] {in-online-news.csv};
\addplot[blue,densely dashed,thick,mark=none] table [x=a, y=c, col sep=comma] {in-online-news.csv};
\addplot[blue,densely dotted,thick,mark=none] table [x=a, y=d, col sep=comma] {in-online-news.csv};
\addplot[green,solid,thick,mark=none] table [x=a, y=e, col sep=comma] {in-online-news.csv};
\addplot[green,densely dashed,thick,mark=none] table [x=a, y=f, col sep=comma] {in-online-news.csv};
\addplot[green,densely dotted,thick,mark=none] table [x=a, y=g, col sep=comma] {in-online-news.csv};
\addplot[orange,solid,thick,mark=none] table [x=a, y=h, col sep=comma] {in-online-news.csv};
\addplot[orange,densely dashed,thick,mark=none] table [x=a, y=i, col sep=comma] {in-online-news.csv};
\addplot[orange,densely dotted,thick,mark=none] table [x=a, y=j, col sep=comma] {in-online-news.csv};
\end{axis}
\end{tikzpicture} &
\begin{tikzpicture}
\begin{axis}[width=.35\textwidth, xmin=0,xmax=117400,xtick={0,25000,50000,75000,100000}, xlabel=TriviaQA,ymin=0,ymax=90,grid=major,font=\small]
\addplot[blue,solid,thick,mark=none] table [x=a, y=b, col sep=comma] {in-online-trv.csv};
\addplot[blue,densely dashed,thick,mark=none] table [x=a, y=c, col sep=comma] {in-online-trv.csv};
\addplot[blue,densely dotted,thick,mark=none] table [x=a, y=d, col sep=comma] {in-online-trv.csv};
\addplot[green,solid,thick,mark=none] table [x=a, y=e, col sep=comma] {in-online-trv.csv};
\addplot[green,densely dashed,thick,mark=none] table [x=a, y=f, col sep=comma] {in-online-trv.csv};
\addplot[green,densely dotted,thick,mark=none] table [x=a, y=g, col sep=comma] {in-online-trv.csv};
\addplot[orange,solid,thick,mark=none] table [x=a, y=h, col sep=comma] {in-online-trv.csv};
\addplot[orange,densely dashed,thick,mark=none] table [x=a, y=i, col sep=comma] {in-online-trv.csv};
\addplot[orange,densely dotted,thick,mark=none] table [x=a, y=j, col sep=comma] {in-online-trv.csv};
\end{axis}
\end{tikzpicture} &
\begin{tikzpicture}
\begin{axis}[width=.35\textwidth, xmin=0,xmax=117400,xtick={0,25000,50000,75000,100000}, xlabel=SearchQA,ymin=0,ymax=90,grid=major,font=\small]
\addplot[blue,solid,thick,mark=none] table [x=a, y=b, col sep=comma] {in-online-sea.csv};
\addplot[blue,densely dashed,thick,mark=none] table [x=a, y=c, col sep=comma] {in-online-sea.csv};
\addplot[blue,densely dotted,thick,mark=none] table [x=a, y=d, col sep=comma] {in-online-sea.csv};
\addplot[green,solid,thick,mark=none] table [x=a, y=e, col sep=comma] {in-online-sea.csv};
\addplot[green,densely dashed,thick,mark=none] table [x=a, y=f, col sep=comma] {in-online-sea.csv};
\addplot[green,densely dotted,thick,mark=none] table [x=a, y=g, col sep=comma] {in-online-sea.csv};
\addplot[orange,solid,thick,mark=none] table [x=a, y=h, col sep=comma] {in-online-sea.csv};
\addplot[orange,densely dashed,thick,mark=none] table [x=a, y=i, col sep=comma] {in-online-sea.csv};
\addplot[orange,densely dotted,thick,mark=none] table [x=a, y=j, col sep=comma] {in-online-sea.csv};
\end{axis}
\end{tikzpicture} \\
\end{tabular} \vspace{-5pt}
\caption[short]{Online in-domain simulation development F1 learning curves. X-axis is the number of examples with feedback observed. ``$x$ w $y$'' denotes initially training with $x$ supervised in-domain examples and simulating with $y$ amount of feedback noise.} 
\label{fig:in-online-lc}
\end{figure*}
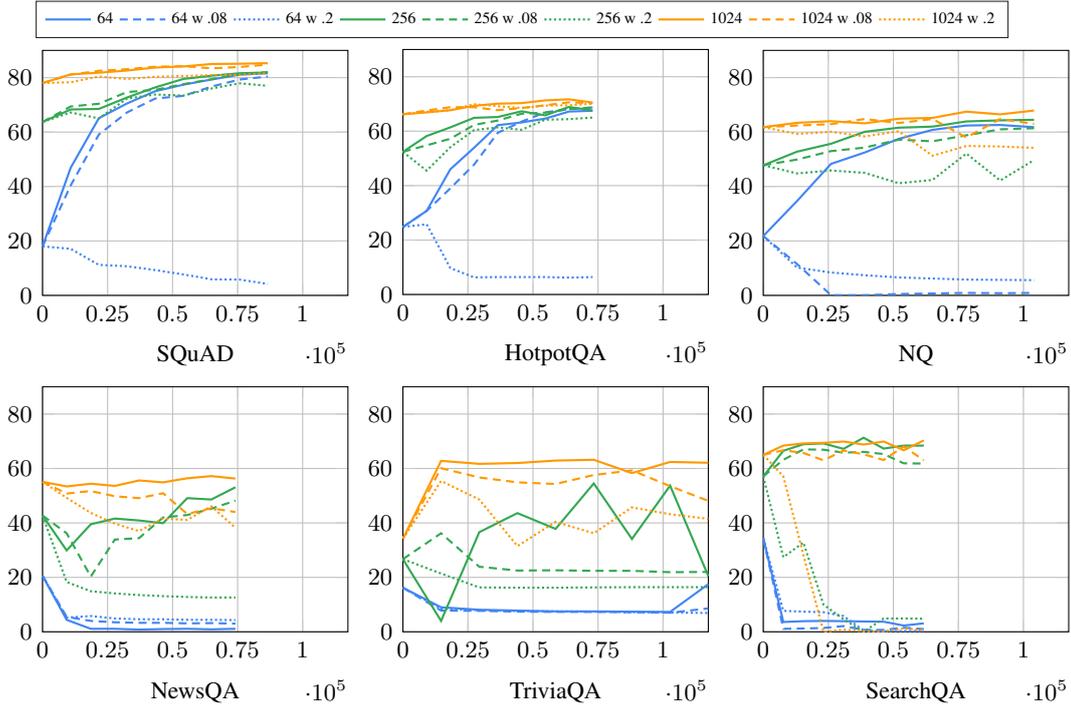

\begin{table*}[ht!]\vspace{-5pt}
\small
    \centering\footnotesize\renewcommand*{\arraystretch}{1.4} 
    \begin{tabular}{>{\bfseries\arraybackslash}l}
        \toprule
        Setup \\ \cmidrule(lr){1-1}
        64+sim\\ 256+sim\\ 1024+sim\\ 
        \bottomrule
    \end{tabular}%
    \csvloop{file=in-pfm-offline.csv, no head, 
        before reading=\centering,
        tabular={*{7}{c}},
        table head=\toprule & \textbf{SQuAD}  & \textbf{HotpotQA}  & \textbf{NQ}  & \textbf{NewsQA}  & \textbf{TriviaQA}  & \textbf{SearchQA}  
        \\\cmidrule(lr){2-2} \cmidrule(lr){3-3} \cmidrule(lr){4-4} \cmidrule(lr){5-5} \cmidrule(lr){6-6} \cmidrule(lr){7-7},
        command=&\csvcoli(\csvcolii) & \csvcoliv(\csvcolv) & \csvcolvii(\csvcolviii) & \csvcolx(\csvcolxi) & \csvcolxiii(\csvcolxiv) & \csvcolxvi(\csvcolxvii), 
        table foot=\bottomrule}
    \vspace{-5pt}
    \caption{Offline in-domain simulation development F1 performance. Numbers in parenthesis show the performance gain (green) or decrease (red) of offline learning compared to online learning (\autoref{fig:in-pfm-online}).}
    \label{tab:in-pfm-offline}
\end{table*}

\begin{table*}[t!]\vspace{-5pt}
    \centering\footnotesize\renewcommand*{\arraystretch}{1.4} 
    \begin{tabular}{>{\bfseries\arraybackslash}l}
        \toprule
        Setup\\ \cmidrule(lr){1-1}
        64+sim\\ 256+sim\\ 1024+sim\\ 
        \bottomrule
    \end{tabular}%
    \csvloop{file=regret.csv, no head, 
        before reading=\centering%
        ,
        tabular={*{7}{c}},
        table head=\toprule & \textbf{SQuAD} & \textbf{HotpotQA} & \textbf{NQ} &
       \textbf{NewsQA} & \textbf{TriviaQA} & \textbf{SearchQA} \\\cmidrule(lr){2-2} \cmidrule(lr){3-3} \cmidrule(lr){4-4} \cmidrule(lr){5-5} \cmidrule(lr){6-6} \cmidrule(lr){7-7},
        command=&\text{\csvcoli}~/~\csvcolii & \csvcoliii~/~\csvcoliv & \csvcolv~/~\csvcolvi & \csvcolvii~/~\csvcolviii & \csvcolix~/~\csvcolx & \csvcolxi~/~\csvcolxii,
        table foot=\bottomrule}
    \vspace{-5pt}
    \caption{Regret averaged by the number of feedback observations in online/offline in-domain simulations. }
    \label{tab:regret}
\end{table*}

\section{Online Learning}\label{sec:online-learning}

We simulate a scenario where only a limited amount of supervised data is available, and the model mainly learns from explicit user feedback on predicted answers. 
We use 64, 256, or 1{,}024 in-domain annotated examples to train an initial model. 
This section focuses on online learning, where the learner updates the model parameters after each feedback is observed (\autoref{alg:online}).

\autoref{fig:in-pfm-online} presents the performance of in-domain simulation with online learning. The performance pattern varies across different datasets. Bandit learning consistently improves performance on \squad, \hotpotqa, and \nqshort across different amounts of supervised data used to train the initial model. The performance gain is larger with weaker initial models (i.e., trained on 64 supervised examples): 63.6 on \squad, 42.7 on \hotpotqa, and 40.0 on \nqshort.
Bandit learning is not always effective on \newsqa, \triviaqa, and \searchqa, especially with weaker initial models. 
This may be attributed to the quality of training set annotations, which determines the accuracy of reward in our setup. \searchqa and \triviaqa use distant supervision to match questions and relevant contexts from the web, likely decreasing reward quality in our setup. While \newsqa is crowdsourced, \citet{Trischler2017NewsQAAM} report relatively low human performance (69.4 F1), possibly indicating data challenges that also decrease our reward quality. %
Learning progression across datasets (\autoref{fig:in-online-lc}) shows that initial models trained with 1{,}024 examples can achieve peak performance with one third or even one quarter of feedback provided.

\paragraph{Feedback Noise Simulation}

\autoref{fig:in-online-lc} shows learning curves with simulated noise via different amounts of feedback perturbation (0\%, 8\%, or 20\%).
When perturbation-free simulation is effective, models remain robust to  noise: 8\% noise results in small fluctuations of the learning curve, but the final performance degrades minimally.
Starting with weaker initial models and learning with a higher noise ratio may cause learning to fail (e.g., simulation on \squad with 64 initial examples and 20\% noise).
When online perturbation-free simulation fails, online learning with noisy feedback fails too.

\paragraph{Sensitivity Analysis}

Training Transformer-based models has been shown to have stability issues, especially when training with limited amount of data~\cite{Zhang2021RevisitingFB}. 
Our non-standard training procedure (i.e., one epoch with a fixed learning rate) may further increase instability. 
We study the stability of the learning process using initial models trained on only 64 in-domain supervised examples on \hotpotqa and \triviaqa: the former shows significant performance gain while the latter shows the opposite. 
We experiment with five initial models trained on different sets of 64 supervised examples, each used to initiate a separate simulation experiment. 
Four out of five experiments on \hotpotqa show performance gains similar to what we observed so far, except one experiment that starts with very low initialization performance. In contrast, nearly all experiments on \triviaqa collapse (mean F1 of 7.3). 
We also conduct sensitivity analysis with stronger initial models trained with 1{,}024 examples, and observe that the final performance is stable across runs on both  \hotpotqa and \triviaqa (standard deviations are 0.5 and 2.6). 
\autoref{tab:sen-anl} in \autoref{sec:extra} provides detailed performance numbers.

\section{Offline Learning} \label{sec:offline-learning}

We simulate offline bandit learning (\autoref{alg:offline}), where feedback is collected all at once with the initial model.
The learning scenario follows the previous section: only a limited amount of supervised data is available (64, 256, or 1{,}024 in-domain examples) to train initial models.

\autoref{tab:in-pfm-offline} shows the performance of offline simulation experiments compared to online simulations. We observe mixed results. On \squad, \hotpotqa, \nqshort, and \newsqa, offline learning outperforms online learning when using stronger initial models (i.e., models trained on 256 and 1{,}024 examples). 
This illustrates the benefit of the more standard training loop, especially with our Transformer-based model that is better optimized with a linear learning rate schedule and multiple epochs, both incompatible with the online setup. 
On \triviaqa and \searchqa, offline simulation is ineffective regardless of the performance of initial models. This result echoes the learning challenges in the online counterparts on these two datasets.

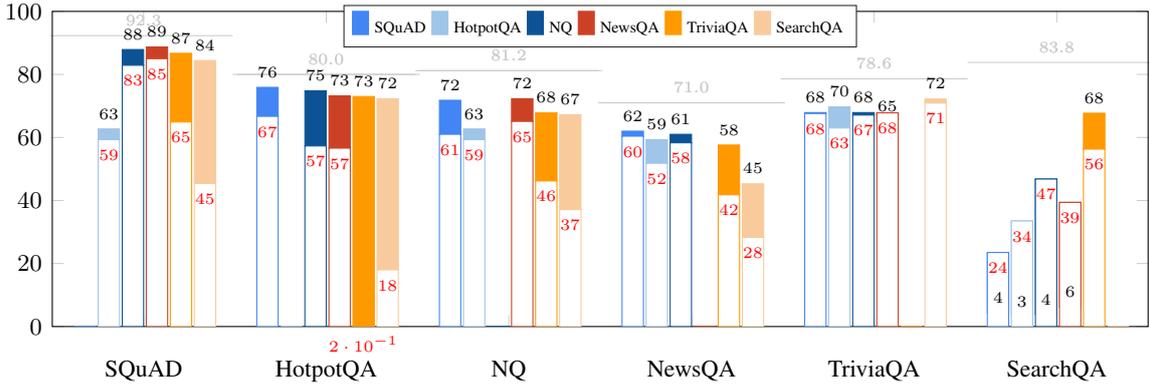
\begin{figure*}[t!]\vspace{-10pt}
\centering
\begin{tikzpicture}[
  /pgfplots/every axis/.style={
    ybar stacked,
    ymin=0,ymax=100,
    x tick label style={yshift=-0.8em,rotate=0,anchor=north},
    symbolic x coords={SQuAD, HotpotQA, NQ, NewsQA, TriviaQA, SearchQA},
  bar width=8,
  width=\textwidth,
  height=.36\textwidth,
  xtick=data, 
  font=\small,
  nodes near coords,
  },
]

\begin{axis}[hide axis, clip=false, xshift=-35, nodes near coords,
    nodes near coords style={font=\tiny, color=gray, /pgf/number format/fixed zerofill, /pgf/number format/precision=1}]
    \addplot [gray, jump mark left, nodes near coords align={anchor=south, xshift=35}] table [x=dataset, y=supervised, col sep=comma] {in-online-pfm.csv};
\end{axis}
\begin{axis}[hide axis, clip=false, xshift=35, nodes near coords, nodes near coords style={font=\tiny, color=gray, /pgf/number format/fixed zerofill, /pgf/number format/precision=1}]
    \addplot [gray, jump mark right, nodes near coords align={anchor=south, xshift=-35}] table [x=dataset, y=supervised, col sep=comma] {in-online-pfm.csv};
\end{axis}

\begin{axis}[bar shift=-22,nodes near coords={\pgfmathfloatifflags{\pgfplotspointmeta}{0}{}{\pgfmathprintnumber{\pgfplotspointmeta}}},nodes near coords style={font=\tiny, xshift=-22, /pgf/number format/fixed zerofill, /pgf/number format/precision=0}]
    \addplot [sqcolor, fill=white, 
    nodes near coords align={anchor=north, color=red}] table [x=dataset, y=sqbase, col sep=comma] {da-onlin-pfm.csv};
    \addplot [sqcolor, fill=sqcolor, nodes near coords align={anchor=south,color=black}, plot graphics/node/.style={text=red}] table [x=dataset, y=sq, col sep=comma] {da-onlin-pfm.csv};
\end{axis}

\begin{axis}[hide axis, bar shift=-13, nodes near coords={\pgfmathfloatifflags{\pgfplotspointmeta}{0}{}{\pgfmathprintnumber{\pgfplotspointmeta}}},nodes near coords style={font=\tiny, xshift=-13, /pgf/number format/fixed zerofill, /pgf/number format/precision=0}]
    \addplot [hpcolor, fill=white, 
    nodes near coords align={anchor=north,color=red}] table [x=dataset, y=hpbase, col sep=comma] {da-onlin-pfm.csv};
    \addplot [hpcolor, fill=hpcolor, nodes near coords align={anchor=south,color=black}] table [x=dataset, y=hp, col sep=comma] {da-onlin-pfm.csv};
\end{axis}

\begin{axis}[hide axis,nodes near coords, bar shift=-4,nodes near coords={\pgfmathfloatifflags{\pgfplotspointmeta}{0}{}{\pgfmathprintnumber{\pgfplotspointmeta}}},
    nodes near coords style={font=\tiny, xshift=-4, /pgf/number format/fixed zerofill, /pgf/number format/precision=0}]
    \addplot [nqcolor, fill=white, 
    nodes near coords align={anchor=north, color=red}] table [x=dataset, y=nqbase, col sep=comma] {da-onlin-pfm.csv};
    \addplot [nqcolor, fill=nqcolor, nodes near coords align={anchor=south,color=black}, plot graphics/node/.style={text=red}] table [x=dataset, y=nq, col sep=comma] {da-onlin-pfm.csv};
\end{axis}

\begin{axis}[hide axis, bar shift=5, nodes near coords={\pgfmathfloatifflags{\pgfplotspointmeta}{0}{}{\pgfmathprintnumber{\pgfplotspointmeta}}},
    nodes near coords style={font=\tiny,xshift=5, /pgf/number format/fixed zerofill, /pgf/number format/precision=0}]
    \addplot [newscolor, fill=white, 
    nodes near coords align={anchor=north,color=red}] table [x=dataset, y=newsbase, col sep=comma] {da-onlin-pfm.csv};
    \addplot [newscolor, fill=newscolor, nodes near coords align={anchor=south,color=black}] table [x=dataset, y=news, col sep=comma] {da-onlin-pfm.csv};
\end{axis}

\begin{axis}[hide axis, bar shift=14, nodes near coords={\pgfmathfloatifflags{\pgfplotspointmeta}{0}{}{\pgfmathprintnumber{\pgfplotspointmeta}}},
    nodes near coords style={font=\tiny,xshift=14, /pgf/number format/fixed zerofill, /pgf/number format/precision=0}]
    \addplot [trvcolor, fill=white, 
    nodes near coords align={anchor=north,color=red}] table [x=dataset, y=trvbase, col sep=comma] {da-onlin-pfm.csv};
    \addplot [trvcolor, fill=trvcolor, nodes near coords align={anchor=south,color=black}] table [x=dataset, y=trv, col sep=comma] {da-onlin-pfm.csv};
\end{axis}

\begin{axis}[hide axis, bar shift=23, nodes near coords={\pgfmathfloatifflags{\pgfplotspointmeta}{0}{}{\pgfmathprintnumber{\pgfplotspointmeta}}},
    nodes near coords style={font=\tiny,xshift=23, /pgf/number format/fixed zerofill, /pgf/number format/precision=0}]
    \addplot [seacolor, fill=white, 
    nodes near coords align={anchor=north,color=red}] table [x=dataset, y=seabase, col sep=comma] {da-onlin-pfm.csv};
    \addplot [seacolor, fill=seacolor, nodes near coords align={anchor=south,color=black}] table [x=dataset, y=sea, col sep=comma] {da-onlin-pfm.csv};
\end{axis}

\begin{axis}[hide axis, xmin=SQuAD,xmax=SearchQA,ymin=0,ymax=100,height=.1\textwidth,
    legend style={draw=white!15!black,at={(0.5,200)},anchor=south},legend columns=-1,font=\tiny,
    ]
        \addlegendimage{sqcolor,fill=sqcolor}
        \addlegendentry{SQuAD};
        \addlegendimage{hpcolor,fill=hpcolor}
        \addlegendentry{HotpotQA};
        \addlegendimage{nqcolor, fill=nqcolor}
        \addlegendentry{NQ};
        \addlegendimage{newscolor, fill=newscolor}
        \addlegendentry{NewsQA};
        \addlegendimage{trvcolor, fill=trvcolor}
        \addlegendentry{TriviaQA};
        \addlegendimage{seacolor,fill=seacolor}
        \addlegendentry{SearchQA};
\end{axis}

\end{tikzpicture}
\caption{Online domain adaptation simulation development F1 performance. Horizontal grey lines represent the supervised training performance on each complete dataset. Bar colors denotes the source domain. Labels in red are the performance of initial models on the target domain (x-axis). Solid colors and black labels represent simulation performance on the target domain. } 
\label{fig:da-pfm-online}
\end{figure*}

\begin{figure*}[t!]\vspace{2pt}
\centering
\begin{tabular}{p{.27\textwidth} p{.27\textwidth} p{.27\textwidth}}
\multicolumn{3}{c}{\begin{tikzpicture} 
    \begin{axis}[hide axis, xmin=10,xmax=50,ymin=0,ymax=0.4,height=.1\textwidth,
    legend style={draw=white!15!black,legend cell align=left},legend columns=-1,font=\tiny,
    ]
        \addlegendimage{sqcolor,solid,thick}
        \addlegendentry{SQuAD};
        \addlegendimage{hpcolor,solid,thick}
        \addlegendentry{HotpotQA};
        \addlegendimage{nqcolor,solid,thick}
        \addlegendentry{NQ};
        \addlegendimage{newscolor,solid,thick}
        \addlegendentry{NewsQA};
        \addlegendimage{trvcolor,solid,thick}
        \addlegendentry{TriviaQA};
        \addlegendimage{seacolor,solid,thick}
        \addlegendentry{SearchQA};
        
    \end{axis}
\end{tikzpicture}} \\
\begin{tikzpicture}
    \begin{axis}[width=.35\textwidth, xmin=0,xmax=117400,xtick={0,25000,50000,75000,100000}, xlabel=SQuAD,ymin=0,ymax=100,grid=major,font=\small]
        \addplot[sqcolor,solid,thick,mark=none] table [x=x, y=sq, col sep=comma] {da-online-sq.csv};
        \addplot[hpcolor,solid,thick,mark=none] table [x=x, y=hp, col sep=comma] {da-online-sq.csv};
        \addplot[nqcolor,solid,thick,mark=none] table [x=x, y=nq, col sep=comma] {da-online-sq.csv};
        \addplot[trvcolor,solid,thick,mark=none] table [x=x, y=trv, col sep=comma] {da-online-sq.csv};
        \addplot[seacolor,solid,thick,mark=none] table [x=x, y=sea, col sep=comma] {da-online-sq.csv};
        \addplot[newscolor,solid,thick,mark=none] table [x=x, y=news, col sep=comma] {da-online-sq.csv};
    \end{axis}
\end{tikzpicture} & 
\begin{tikzpicture}
    \begin{axis}[width=.35\textwidth, xmin=0,xmax=117400,xtick={0,25000,50000,75000,100000}, xlabel=HotpotQA,ymin=0,ymax=100,grid=major,font=\small]
        \addplot[sqcolor,solid,thick,mark=none] table [x=x, y=sq, col sep=comma] {da-online-hp.csv};
        \addplot[hpcolor,solid,thick,mark=none] table [x=x, y=hp, col sep=comma] {da-online-hp.csv};
        \addplot[nqcolor,solid,thick,mark=none] table [x=x, y=nq, col sep=comma] {da-online-hp.csv};
        \addplot[trvcolor,solid,thick,mark=none] table [x=x, y=trv, col sep=comma] {da-online-hp.csv};
        \addplot[seacolor,solid,thick,mark=none] table [x=x, y=sea, col sep=comma] {da-online-hp.csv};
        \addplot[newscolor,solid,thick,mark=none] table [x=x, y=news, col sep=comma] {da-online-hp.csv};
    \end{axis}
\end{tikzpicture} & 
\begin{tikzpicture}
    \begin{axis}[width=.35\textwidth, xmin=0,xmax=117400,xtick={0,25000,50000,75000,100000}, xlabel=NQ,ymin=0,ymax=100,grid=major,font=\small]
        \addplot[sqcolor,solid,thick,mark=none] table [x=x, y=sq, col sep=comma] {da-online-nq.csv};
        \addplot[hpcolor,solid,thick,mark=none] table [x=x, y=hp, col sep=comma] {da-online-nq.csv};
        \addplot[nqcolor,solid,thick,mark=none] table [x=x, y=nq, col sep=comma] {da-online-nq.csv};
        \addplot[trvcolor,solid,thick,mark=none] table [x=x, y=trv, col sep=comma] {da-online-nq.csv};
        \addplot[seacolor,solid,thick,mark=none] table [x=x, y=sea, col sep=comma] {da-online-nq.csv};
        \addplot[newscolor,solid,thick,mark=none] table [x=x, y=news, col sep=comma] {da-online-nq.csv};
    \end{axis}
\end{tikzpicture} \\
\begin{tikzpicture}
    \begin{axis}[width=.35\textwidth, xmin=0,xmax=117400,xtick={0,25000,50000,75000,100000}, xlabel=NewsQA,ymin=0,ymax=100,grid=major,font=\small]
        \addplot[sqcolor,solid,thick,mark=none] table [x=x, y=sq, col sep=comma] {da-online-news.csv};
        \addplot[hpcolor,solid,thick,mark=none] table [x=x, y=hp, col sep=comma] {da-online-news.csv};
        \addplot[nqcolor,solid,thick,mark=none] table [x=x, y=nq, col sep=comma] {da-online-news.csv};
        \addplot[trvcolor,solid,thick,mark=none] table [x=x, y=trv, col sep=comma] {da-online-news.csv};
        \addplot[seacolor,solid,thick,mark=none] table [x=x, y=sea, col sep=comma] {da-online-news.csv};
        \addplot[newscolor,solid,thick,mark=none] table [x=x, y=news, col sep=comma] {da-online-news.csv};
    \end{axis}
\end{tikzpicture} & 
\begin{tikzpicture}
    \begin{axis}[width=.35\textwidth, xmin=0,xmax=117400,xtick={0,25000,50000,75000,100000}, xlabel=TriviaQA,ymin=0,ymax=100,grid=major,font=\small]
        \addplot[sqcolor,solid,thick,mark=none] table [x=x, y=sq, col sep=comma] {da-online-trv.csv};
        \addplot[hpcolor,solid,thick,mark=none] table [x=x, y=hp, col sep=comma] {da-online-trv.csv};
        \addplot[nqcolor,solid,thick,mark=none] table [x=x, y=nq, col sep=comma] {da-online-trv.csv};
        \addplot[trvcolor,solid,thick,mark=none] table [x=x, y=trv, col sep=comma] {da-online-trv.csv};
        \addplot[seacolor,solid,thick,mark=none] table [x=x, y=sea, col sep=comma] {da-online-trv.csv};
        \addplot[newscolor,solid,thick,mark=none] table [x=x, y=news, col sep=comma] {da-online-trv.csv};
    \end{axis}
\end{tikzpicture} &
\begin{tikzpicture}
    \begin{axis}[width=.35\textwidth, xmin=0,xmax=117400,xtick={0,25000,50000,75000,100000}, xlabel=SearchQA,ymin=0,ymax=100,grid=major,font=\small]
        \addplot[sqcolor,solid,thick,mark=none] table [x=x, y=sq, col sep=comma] {da-online-sea.csv};
        \addplot[hpcolor,solid,thick,mark=none] table [x=x, y=hp, col sep=comma] {da-online-sea.csv};
        \addplot[nqcolor,solid,thick,mark=none] table [x=x, y=nq, col sep=comma] {da-online-sea.csv};
        \addplot[trvcolor,solid,thick,mark=none] table [x=x, y=trv, col sep=comma] {da-online-sea.csv};
        \addplot[seacolor,solid,thick,mark=none] table [x=x, y=sea, col sep=comma] {da-online-sea.csv};
        \addplot[newscolor,solid,thick,mark=none] table [x=x, y=news, col sep=comma] {da-online-sea.csv};
    \end{axis}
\end{tikzpicture}\\
\end{tabular}
\caption[short]{Online domain adaptation simulation development F1 learning curves. X-axis is the number of examples with feedback observed. Colors denote the source domain.}
\label{fig:da-online-lc}
\end{figure*}
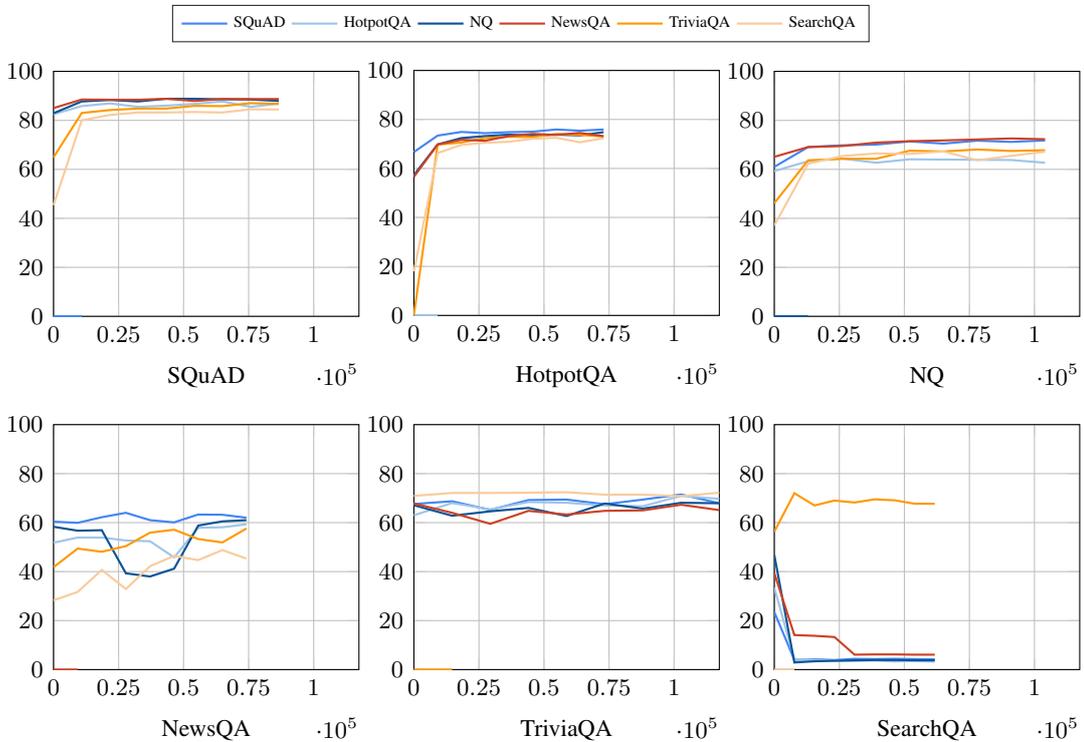

\paragraph{Online vs. Offline Regret}

\autoref{tab:regret} compares online and offline regret. Regret numbers are averaged over the number of feedback observations.\footnote{\autoref{tab:perc-pos-online-in} in \autoref{sec:extra} lists the percentage of positive feedback in online and offline in-domain simulation.} 
Online learning generally displays lower regret for similar initial models on \squad, \hotpotqa, and \nqshort. This is expected because later interactions in the simulation can benefit from early feedback in online learning. In contrast, in our offline scenario, we only update after seeing all examples, so regret numbers depend on the initial model only. 
Regret results on \newsqa, \triviaqa, and \searchqa are counterintuitive, generally showing that online learning has similar or higher regret. 
The cases showing significantly higher online regret (64+sim on \newsqa and \searchqa) can be explained by the learning failing, which impacts online regret, but not our offline regret. 
The others are more complex, and we hypothesize that they may be because of combination of (a) inherent noise in the data; and (b) in cases where online learning is effective, the gap between the strictly-defined reward that is used to compute regret and the relaxed F1 evaluation metric.  
Further analysis is required for a more conclusive conclusion.

\section{Domain Adaptation}\label{sec:domain-adaptation}

Learning from user feedback creates a compelling avenue to deploy systems that target new domains not addressed by existing datasets. 
The scenario we simulate in this section starts with training a QA model on a complete existing annotated dataset, and deploying it to interact with users and learn from their feedback in a new domain. 
We do not assume access to any annotated training data in the target domain. 
We report experiments with online learning. Offline adaptation experiments are discussed in Appendix~\ref{sec:offline-adaptation}.

\autoref{fig:da-pfm-online} shows online domain adaptation performance. 
On 22/30 configurations, online adaptation introduces significant performance gains ($>$2 F1 score). For example, adapting from \triviaqa and \searchqa to the other four domains improves performance by 27--72.8 F1. On \hotpotqa, the model initially trained on \triviaqa shows an impressive adaptation, improving from 0.2 F1 to 73 F1.\footnote{We replicate this result with different model initializations to confirm it is not random.}

Our simulations show reduced effectiveness when the target domain is either \triviaqa or \searchqa, likely because the simulated feedback is based on noisy distantly supervised data.  
For \searchqa, the low performance of initial models from other domains may also contribute to the adaptation failure. As expected, this indicates the effectiveness of the process depends on the relation between the source and target domains. 
\searchqa seems farthest from the other domains, mirroring observations from prior work~\cite{Su2019GeneralizingQA}.

\autoref{fig:da-online-lc} shows learning curves for our simulation experiments. 
Generally, we observe the choice of source and target domains influences adaptation rates.
Models quickly adapt to \squad, \hotpotqa, and \nqshort, reaching near final performance with a quarter of the total feedback provided. On \newsqa, models initially trained on \triviaqa and \searchqa adapt slower than those initially trained on other three datasets.
On \triviaqa, we observe little change in performance throughout simulation. 
On \searchqa, only the model initially trained on \triviaqa shows a performance gain. Both \searchqa and  \triviaqa include context paragraphs from the web, potentially making domain adaptation from one to the other easier.

Lastly, we compare bandit learning with initial models trained on a small amount of in-domain data (\autoref{sec:online-learning}) and initial models trained on a large amount of out-of-domain data. 
\autoref{tab:comparison} compares online learning with initial models trained on 1{,}024 in-domain supervised examples and online domain adaptation with a \squad-initialized model. %
\squad initialization provides a robust starting point for all datasets except \searchqa. On four out of five datasets, the final performance is better with \squad-initialized model. This is potentially because the model is exposed to different signals from two datasets and overall sees more data, either as supervised examples or through feedback. However, on \searchqa, learning with \squad-initialized model performs much worse than learning with the initial model trained on 1{,}024 in-domain examples, potentially because of the gap in initial model performance (23.5 vs. 65 F1). %

\begin{table}[t]
\centering\footnotesize
\begin{tabular}{l c c}
\toprule
\textbf{Dataset} & \textbf{In-domain} & \textbf{\squad-initialized} \\
\midrule
\hotpotqa & 66.2 $\rightarrow$ 70.5 & \textbf{66.7} $\rightarrow$ \textbf{75.9} \\
\nqshort  & \textbf{61.8} $\rightarrow$ 67.9 & 61.0 $\rightarrow$ \textbf{71.8} \\
\newsqa & 55.1 $\rightarrow$ 56.3 & \textbf{60.4} $\rightarrow$ \textbf{62.0} \\
\triviaqa &  34.2 $\rightarrow$ 62.1 & \textbf{67.6} $\rightarrow$ \textbf{67.9}  \\
\searchqa &  \textbf{65.0} $\rightarrow$ \textbf{70.3} & 23.5 $\rightarrow$ \space\space 4.2 \\
\bottomrule
\end{tabular}
\caption{\label{tab:comparison}
Online learning development F1 Comparison between in-domain with initial models trained on 1{,}024 supervised examples, and adaptation with \squad as the source domain. Each entry provide performance before (right-side of arrow) and after (left-side) feedback simulation. Higher before/after performance is in bold. }\vspace{-3pt}
\end{table}
\pdfoutput=1
\section{Related Work}\label{sec:related}

Bandit learning has been applied to a variety of NLP problems including neural machine translation~\cite[NMT;][]{Sokolov2017AST,Kreutzer2018CanNM, Kreutzer2018ReliabilityAL, Mendoncca2021OnlineLM}, structured prediction~\cite{Sokolov2016LearningSP}, semantic parsing~\cite{Lawrence2018:sempar-human-feedback}, intent recognition~\cite{Falke2021:slu-feedback}, and summarization~\cite{gunasekara-etal-2021-using-question}. 
Explicit human feedback has been studied as a direct learning signal for NMT~\cite{Kreutzer2018ReliabilityAL, Mendoncca2021OnlineLM}, semantic parsing~\cite{artzi-zettlemoyer:2011:EMNLP,Lawrence2018:sempar-human-feedback}, and summarization~\cite{Stiennon2020LearningTS}. 
\citet{Nguyen2017ReinforcementLF} simulates bandit feedback to improve an MT system fully trained on a large annotated dataset, including analyzing robustness to feedback perturbations.
Our work shows that simulated bandit feedback is an effective learning signal for extractive question answering tasks. 
Our work differs in focus on reducing annotation costs by relying on few annotated examples only to train the initial model, or by eliminating the need for in-domain annotation completely by relying on data in other domains to train initial models. 
Implicit human feedback, where feedback is derived from human behavior rather than explicitly requested, has also been studied, including for dialogue~\cite{Jaques2020:human-centric-dialog} and instruction generation~\cite{Kojima2021:gen-learn}. 
We focus on explicit feedback, but implicit signals also hold promise to improve QA systems.

Alternative forms of supervision for QA have been explored in prior work, such as explicitly providing fine-grained information~\cite{Dua2020BenefitsOI, Khashabi2020MoreBF}. \citet{Kratzwald2020LearningAC} resembles our setting in seeking binary feedback to replace span annotation, but their goal is to create supervised data more economically. \citet{Campos2020ImprovingCQ} proposes feedback-weighted learning to improves conversational QA using simulated binary feedback. Their approach relies on multiple samples (i.e., feedback signals) per example, training for multiple epochs online by re-visiting the same questions repeatedly, and tuning two additional hyperparameters. In contrast, we study improving QA systems via feedback as a bandit learning problem. In both online and offline setups, we assume only one feedback sample per example. We also provide extensive sensitivity studies to the amount of annotations available, different model initialization, and noisy feedback across various datasets.

Domain adaptation for QA has been widely studied~\cite{fisch2019mrqa,Khashabi2020UnifiedQACF}, including using data augmentation~\cite{Yue2021ContrastiveDA}, adversarial training~\cite{Lee2019DomainagnosticQW}, contrastive method~\cite{Yue2021ContrastiveDA}, back-training~\cite{Kulshreshtha2021BackTrainingES}, and exploiting small lottery subnetworks~\cite{zhu-etal-2021-less-domain}.

\section{Conclusion}\label{sec:disc}

We present a simulation study of learning from user feedback for extractive QA. We formulate the problem as contextual bandit learning. 
We conduct experiments to show the effectiveness of such feedback, the robustness to feedback noise, the impact of initial model performance, the trade-offs between online and offline learning, and the potential for domain adaptation. 
Our study design emphasizes the potential for reducing annotation costs by annotating few examples or by utilizing existing datasets for new domains. 

We intentionally adopt a basic setup, including a simple binary reward and vanilla learning algorithms, to illustrate what can be achieved with a relatively simple variant of the contextual bandit learning scenario. 
Our results already indicate the strong potential of learning from feedback, which more advanced methods are likely to further improve. 
For example, the balance between online and offline learning can be further explored using proximal policy optimization~\cite[PPO;][]{Schulman2017:ppo} or replay memory~\cite{Mnih:15humanlevelatari}. 
With well-designed interface, human users may be able to provide more sophisticated feedback~\cite{Lamm2021QEDAF}, which will provide a stronger signal compared to our binary reward. 

Our aim in this study is to lay the foundation for future work, by formalizing the setup and showing its potential. 
This is a critical step in enabling future research, especially going beyond simulation to study using \emph{real} human feedback for QA systems. 
Another important direction for future work is studying user feedback for QA systems that do both context retrieval and answer generation~\cite{Lewis2020RetrievalAugmentedGF}, where assigning the feedback to the appropriate stage in the process poses a challenge. 
Beyond extractive QA, we hope our work will inspire research of user feedback as a signal to improve other types of NLP systems.

\eat{
\ggao{ 
an alternative version of discussion below: not only mention the study but also the conclusion (could use this longer version if space allowed) unfinished draft}
\begin{itemize}
    \item \textbf{Offline Adaptation} We find that online learning is generally more effective on domain adaptation, but we observe that offline adaption performs slightly better when both domains are collected from Wikipedia. (\autoref{tab:da-pfm-offline})
    \item \textbf{Lightweight Model} We find that updating the entire model via bandit learning performs better than only updating a small portion of the model parameters, even though the latter is more common in RL literature.
    \item \textbf{Online and Offline Combination} \td{add a paragraph in appendix}
    \item \textbf{Partial Reward} We study F1 rewards and more complicated reward perturbations, which show similar outcome to our default setup.
\end{itemize}

}

\section*{Legal and Ethical Considerations}

Our work's limitations are discussed in \autoref{sec:intro} and \autoref{sec:disc}.
All six datasets we use are from prior work, are publicly available, and are commonly used for the study of extractive QA. \autoref{sec:exp-setup} reports our computational budget and experimental setup in detail. Our codebase is available at \url{https://github.com/lil-lab/bandit-qa}.

\section*{Acknowledgements}

This research was supported by ARO W911NF-21-1-0106, NSF under grants No. 1750499, the NSF AI Institute for the Foundations of Machine Learning (IFML), and a Google Faculty Research Award. 
Finally, we thank the action editor and the anonymous reviewers for detailed comments.

\bibliography{custom, mainref}
\bibliographystyle{acl_natbib}

\clearpage

\appendix

\section{Additional Discussion}
\label{sec:discussion}

\paragraph{Reward Function} 
Intuitively, partial credit reward  may improve learning over binary rewards. We experiment with using F1 score of the predicted answer span as a more refined feedback.\footnote{We set the reward as -0.1 if receiving a 0 F1 score. In general, updating with negative rewards consistently shows a slightly higher performance across different setups for both binary and F1 reward.} In practice, this does not introduce a stronger learning signal, potentially because the distribution over F1 scores is bimodal and focused on extreme values: around 85 \% F1 scores are either 0 or 1 for predicted spans from a \squad-trained model on 8\% NQ training data. We observe similar trends on all six datasets across all setups. 
Experiments with BLEU score~\cite{Papineni2002BleuAM} as feedback show similar conclusion and distribution to F1 score.

\paragraph{Perturbation} 
In practice, noise in feedback is likely to be more systematic than the statistical simplification which defines noise as the random percentage of wrong feedback. For example, prior work~\cite{Nguyen2017ReinforcementLF} on bandit neural machine translation (NMT) proposes that noisy human feedback is granular, high-variance, and skewed, which can be approximated by mathematical functions and shows to significantly impact the bandit NMT learning.
We experiment with the three perturbation functions from \citet{Nguyen2017ReinforcementLF} on F1 reward. Our experiments show that the effect of adding these perturbation functions is negligible. We hypothesize that the reward distribution for NMT is likely to be closer to a normal distribution, rather than a bimodal one like QA.

\section{Additional Experiments}\label{sec:extra}

\subsection{Method of Sampling}\label{sec:sampling}
\begin{table}[t!]
\centering\small
\begin{tabular}{l r r}
\toprule
\textbf{Dataset} & $\mathbf{\arg\max}$ & \textbf{Sampling} \\
\midrule
\squad & 80.0 & 73.6 \\
\hotpotqa & 65.7 & 56.8 \\
\nqshort  & 64.8 & 62.9\\
\bottomrule
\end{tabular}
\caption{\label{tab:sampling}
Comparison of final F1 development scores between $\arg\max$ and sampling in online simulation with initial models trained on 256 supervised in-domain examples.}
\end{table}
While $\arg\max$ can bias towards exploitation, sampling can encourage  more exploration.
We experiment with prediction via $\arg\max$ and sampling from the output distribution over spans. \autoref{tab:sampling} shows that $\arg\max$ performs better than random sampling on three datasets. This set of experiments is conducted with batch size 80. 

\subsection{Sensitivity Analysis}\label{sec:sen-anl}

\autoref{tab:sen-anl} shows the sensitivity analysis results for online in-domain simulation on \hotpotqa and \triviaqa. 
We experiment with five initial models trained on different sets of 64 or 1{,}024 supervised examples, each used to initiate a separate simulation experiment. For weaker initial models trained on 64 supervised examples, four out of five experiments on \hotpotqa show performance gains similar to our main results, except one experiment that starts with a very low initialization performance. Nearly all experiments on \triviaqa collapse (mean F1 of 7.3). Our sensitivity analysis with stronger initial models trained on 1{,}024 examples shows that the final performance is stable across runs on both \hotpotqa and \triviaqa (standard deviations are 0.5 and 2.6). 

\begin{table*}[t!]
\centering\footnotesize
\begin{tabular}{c|c c |c c}
\toprule
\textbf{Setup} & \multicolumn{2}{c}{\textbf{64 + sim}} & \multicolumn{2}{c}{\textbf{1{,}024 + sim}} \\
& \textbf{HotpotQA} & \textbf{TriviaQA} & \textbf{HotpotQA} & \textbf{TriviaQA} \\
\midrule
42 & 16.4 $\rightarrow$ 66.8 & 16.6 $\rightarrow$ 3.3 & 66.1 $\rightarrow$ 71.5 & 55.1 $\rightarrow$ 58.9 \\
43  & 15.9 $\rightarrow$ 69.7 & 24.0 $\rightarrow$ 3.4 & 65.3 $\rightarrow$ 71.6 & 63.0 $\rightarrow$ 65.0\\
44 & 18.1 $\rightarrow$ 68.8 & 23.3 $\rightarrow$ 2.4 & 66.4 $\rightarrow$ 71.3 & 58.0 $\rightarrow$ 65.1 \\
45 &  6.7 $\rightarrow$ 1.4 & 22.8 $\rightarrow$ 9.9 & 65.1 $\rightarrow$ 71.9 & 60.8 $\rightarrow$ 64.2 \\
46 &  24.8 $\rightarrow$ 67.5 & 16.2 $\rightarrow$ 17.4 & 66.2 $\rightarrow$ 70.5 & 34.2 $\rightarrow$ 62.1 \\
\midrule
$\mu_{\sigma}$ & 16.4\textsubscript{6.5} $\rightarrow$ 54.8\textsubscript{29.9} & 20.6\textsubscript{3.8} $\rightarrow$ 7.3\textsubscript{6.4} & 65.8\textsubscript{0.6} $\rightarrow$ 71.4\textsubscript{0.5} & 54.0\textsubscript{11.4} $\rightarrow$ 63.1\textsubscript{2.6} \\
\bottomrule
\end{tabular}
\caption{
Sensitivity analysis: development F1 scores of online in-domain simulation on \hotpotqa and \triviaqa with initial models trained on 64 or 1{,}024 examples. Each row corresponds to a different random seed and a different set of initial model training examples. $x \rightarrow y$ denotes that the performance changes from $x$ to $y$ after the model learns from feedback. Bottom row reports the mean and standard deviation across the five runs.}\label{tab:sen-anl}
\end{table*}

\subsection{Offline Adaptation}\label{sec:offline-adaptation}

\begin{table*}[t!]
    \centering
    \small
    \csvreader[tabular=lcccccc,
        table head=\toprule Sim+Eval\textbackslash\textbf{Pre-Train} & \textbf{SQuAD} & \textbf{HotpotQA} & \textbf{NQ} & \textbf{NewsQA} & \textbf{TriviaQA} & \textbf{SearchQA} \\\midrule,
        late after line=\\, late after last line=\\\bottomrule]%
        {da-pfm-offline-row.csv}{num=\num, sq=\sq, nq=\nq,hp=\hp,news=\news,trv=\trv,sea=\sea}%
        {\num & \sq & \hp & \nq & \news & \trv & \sea}
    \caption{Offline domain adaptation simulation development F1 performance. Numbers in parenthesis show the performance gain (green) or decrease (red) of offline learning compared to online learning (\autoref{fig:da-pfm-online}). We omit offline adaptation to \searchqa because of our previous observation that all online adaptations to \searchqa fail. }
    \label{tab:da-pfm-offline}
\end{table*}
We perform domain adaptation with offline learning, and compare its performance with online adaptation. \autoref{tab:da-pfm-offline} shows the performance gain of offline adaptation simulation compared to the online setup. 
In most settings, online learning proves to be more effective, possibly because it observes feedback from partially adapted model predictions. In a few settings (4/30), we observe better adaptation with offline settings (+1.1 to +4.6).
Overall, we observe that online learning is more effective on domain adaptation, while offline adaption performs slightly better when both domains are related (e.g., same source domain).

\begin{table*}
\centering
\small
\begin{tabular}{lrrrrr}
\toprule
\textbf{Dataset} & \textbf{Train} & \textbf{Dev} & \textbf{Question (Q)} & \textbf{Context (C)} &  \textbf{Q $\independent$ C} \\ %
\midrule
SQuAD   & 86,588   & 10,507 & Crowdsourced      & Wikipedia & \xmark \\ %
HotpotQA   & 72,928      & 5,904 & Crowdsourced      & Wikipedia & \xmark \\ %
NQ   & 104,071   & 12,836  & Search logs  & Wikipedia   & \cmark  \\ %
NewsQA   & 74,160 & 4,212 & Crowdsourced  & News articles & \cmark  \\ %
TriviaQA$^\spadesuit$     & 61,688      & 7,785      & Trivia  & Web snippets & \cmark  \\ %
SearchQA$^\spadesuit$   & 117,384  & 16,980   & Jeopardy   & Web snippets & \cmark  \\ %
\bottomrule
\end{tabular}
\caption{Dataset statistics. $\spadesuit$-marked datasets use distant supervision to match questions and contexts. $Q \independent C$ is true if the question was written independently from the passage used for context.}
\label{tab:data}
\end{table*}

\begin{table*}[t!]
    \centering\footnotesize\renewcommand*{\arraystretch}{1.4} 
    \begin{tabular}{>{\bfseries\arraybackslash}l}
        \toprule
        Setup\\ \cmidrule(lr){1-1}
        64+sim\\ 256+sim\\ 1024+sim\\ 
        \bottomrule
    \end{tabular}%
    \csvloop{file=perc_pos.csv, no head, 
        before reading=\centering%
        ,
        tabular={*{7}{c}},
        table head=\toprule & \textbf{SQuAD} & \textbf{HotpotQA} & \textbf{NQ} &
       \textbf{NewsQA} & \textbf{TriviaQA} & \textbf{SearchQA} \\\cmidrule(lr){2-2} \cmidrule(lr){3-3} \cmidrule(lr){4-4} \cmidrule(lr){5-5} \cmidrule(lr){6-6} \cmidrule(lr){7-7},
        command=&\text{\csvcoli}/\csvcolii & \csvcoliii/\csvcoliv & \csvcolv/\csvcolvi & \csvcolvii/\csvcolviii & \csvcolix/\csvcolx & \csvcolxi/\csvcolxii,
        table foot=\bottomrule}
    \caption{Percentage of positive examples in online/offline in-domain simulation in one pass on the training set.}
    \label{tab:perc-pos-online-in}
\end{table*}

\end{document}